\def\year{2022}\relax
\documentclass[letterpaper]{article} 
\usepackage{aaai22}  
\usepackage{times}  
\usepackage{helvet}  
\usepackage{courier}  
\usepackage[hang,flushmargin]{footmisc}
\usepackage[hyphens]{url}  
\usepackage{graphicx} 
\usepackage{natbib}  
\usepackage{caption} 
\DeclareCaptionStyle{ruled}{labelfont=normalfont,labelsep=colon,strut=off} 
\usepackage{algorithm}
\usepackage{algorithmic}

\usepackage{amssymb} 
\usepackage{amsmath} 
\usepackage{array} 
\newcolumntype{P}[1]{>{\centering\arraybackslash}p{#1}} 
\usepackage{booktabs} 
\usepackage{multirow} 
\usepackage{subfigure} 
\usepackage{paralist}

\usepackage{newfloat}
\usepackage{listings}
\floatstyle{ruled}
\newfloat{listing}{tb}{lst}{}
\floatname{listing}{Listing}
\title{Powerful Graph Convolutioal Networks with Adaptive Propagation Mechanism for Homophily and Heterophily}
\author {
    Tao Wang\textsuperscript{\rm{1},}\setcounter{footnote}{1}\footnote{Both authors contributed equally to this research},
    Rui Wang\textsuperscript{\rm{1},}\footnotemark[2],
    Di Jin\textsuperscript{\rm{1}},
    Dongxiao He\textsuperscript{\rm{1},}\setcounter{footnote}{0}\footnote{Corresponding author} ,
    Yuxiao Huang\textsuperscript{\rm{2}}
}
\affiliations {
    \textsuperscript{\rm{1}} College of Intelligence and Computing, Tianjin University, Tianjin, China\\
    \textsuperscript{\rm{2}}  Data Scienc, George Washington University, Washington, D.C., USA\\
    \{2019216113, wr1895, jindi, hedongxiao\}@tju.edu.cn, yuxiaohuang@gwu.edu
}

\usepackage{bibentry}

\begin{document}

\maketitle

\begin{abstract}
Graph Convolutional Networks (GCNs) have been widely applied in various fields due to their significant power on processing graph-structured data. Typical GCN and its variants work under a homophily assumption (i.e., nodes with same class are prone to connect to each other), while ignoring the heterophily which exists in many real-world networks (i.e., nodes with different classes tend to form edges). Existing methods deal with heterophily by mainly aggregating higher-order neighborhoods or combing the immediate representations, which leads to noise and irrelevant information in the result. But these methods did not change the propagation mechanism which works under homophily assumption (that is a fundamental part of GCNs). This makes it difficult to distinguish the representation of nodes from different classes. To address this problem, in this paper we design a novel propagation mechanism, which can automatically change the propagation and aggregation process according to homophily or heterophily between node pairs. To adaptively learn the propagation process, we introduce two measurements of homophily degree between node pairs, which is learned based on topological and attribute information, respectively. Then we incorporate the learnable homophily degree into the graph convolution framework, which is trained in an end-to-end schema, enabling it to go beyond the assumption of homophily. More importantly, we theoretically prove that our model can constrain the similarity of representations between nodes according to their homophily degree. Experiments on seven real-world datasets demonstrate that this new approach outperforms the state-of-the-art methods under heterophily or low homophily, and gains competitive performance under homophily.
\end{abstract}

\section{Introduction}
Networks (such as social networks, citation networks, biological networks and traffic networks) are ubiquitous structures that can model relational data. Network analysis \cite{networkanalysis1, networkanalysis2, cao, survey,ArmGAN} has been a hot research topic for decades and has been widely used in many scientific fields such as computer science, social science, biology and physics \cite{socialscience, biology}. Recently, graph convolutional network (GCN) \cite{GCN}, which exhibits significant power on processing graph-structured data, has gained great success and already been adapted in various network analysis tasks, including node classification, community detection, anomaly detection and recommender system \cite{NodeClassification, CommunityDetection, LinkPrediction, anomalydetection, Recommendation, WWW}.

Although many GCN-based methods have been proposed in recent years, such as GraphSage \cite{GraphSage}, GAT \cite{GAT}, MixHop \cite{MixHop}, and HIN \cite{HIN}, they implicitly assume that most connected nodes are from the same class or with similar attributes, which is typically called the homophily of network structure. GCN-based methods reflect the homophily assumption by feature propagation and aggregation within graph neighborhoods. And these methods have shown satisfactory performance in many downstream tasks on network with homophily. However, in real world, there are also many networks where most connected nodes are from different classes, which is called heterophily or low homophily. For example, in dating networks, most people tend to connect with people of the opposite gender. In protein networks, different types of amino are more likely to connect with each other. Under heterophily or low homophily, these methods suffer from poor performance, as the propagation mechanism within graph neighborhoods, which is the most fundamental part of GCN, is problematic and will mix irrelevant information from different classes. Therefore, existing GCN-based methods cannot adapt to the scenario of heterophily or low homophily.

Recently, some efforts have been dedicated to generalizing GCN to heterophilic networks. For example, Geom-GCN \cite{Geom-GCN} proposes a novel geometric aggregation scheme which aggregates immediate neighborhoods and distant nodes that have a certain similarity with the target node in a continuous space. H2GCN \cite{H2GCN} applies some key designs, such as higher-order neighborhoods aggregation and combination of intermediate representations, to boost learning from graph with heterophily. GPR-GNN \cite{GPR-GNN} deals with heterophily and oversmoothing by combining each step of feature propagation with a learnable weight. GGCN \cite{GGCN} allows negative message propagation between graph neighborhoods based on the similarity of representations in order to decouple the heterophily and oversmoothing problems. CPGNN \cite{CPGNN} incorporates an interpretable compatibility matrix for modeling the heterophily or homophily level in graphs, enabling it to go beyond the assumption of strong homophily. However, these methods suffer from serious problems. They coarsely aggregate higher-order (distant neighbors), or combine the intermediate representations, to deal with heterophily. While doing so can fuse effective information to some extent, it will also result in introducing noise and irrelevant information that influence negatively the prediction of downstream tasks. Most importantly, these methods do not change the propagation mechanism which is the essential part of GCN and is problematic under heterophily. This also make the representation of nodes from different classes mixed and indistinguishable.

To solve this problem, we focus on designing an adaptive propagation mechanism for both heterophilic and homophilic networks, and giving a new \textbf{HO}mophily-\textbf{G}uided \textbf{G}raph \textbf{C}onvolutional \textbf{N}etwork called HOG-GCN. In this new approach, we introduce a homophily degree matrix into the graph convolution framework, which is used to model the homophily and heterophily of networks and further conduct the propagation process. This homophily degree matrix can be learned from the attribute and topology information via extracting class-aware information during the propagation process. As a result, the new graph convolution framework can automatically change the feature propagation process via modeling the homophily degree between node pairs using homophily degree matrix. The learning process of homophily degree matrix and the feature propagation process are trained jointly in an end-to-end fashion. Finally, we theoretically prove that the adaptive propagation process can constrain the similarity of representations between nodes according to homophily degree.

\section{Preliminaries}
We first give the notations and problem descriptions, then give the definition of homophily ratio. 
\subsection{Notations and Problem Descriptions}

Given an undirected, unweighted and attributed network $G = (V, E, X)$, where $V = \left \{v_1,v_2, . . . ,v_n \right \}$ is a set of $n$ nodes,  $E = \left \{e_{ij} \right \} \subseteq V  \times V$ is a set of edges, and $X \in \mathbb{R}^{n \times f}$ is a set of node attributes, where $m$ represents the number of attributes. The $i$-th row of $X$ represents the attributes of node $v_i$. The topological structure of $G$ is represented by an adjacency matrix $A=[a_{ij}] \in \mathbb{R}^{n \times n}$, where $a_{ij}=1$ if nodes $v_i$ and $v_j$ are connected, or $a_{ij} = 0$ otherwise. We focus on the semi-supervised node classification task in this paper. This is, assume each node belongs to one out of $C$ classes and we have known the labels of a set of nodes $V_L$ with $|V_L| \ll n$. Each node $v_i \in V_L$ is assigned to a label $y_i \in L = \left \{1, 2,...C\right \}$. The objective of node classification task is to predict the labels of $V\backslash {V_L}$.

\noindent \emph{Definition 1}. \textbf{Homophily Ratio.} Given a network $G = (V, E, X)$, the homophily ratio $h = \frac{{|\{ (u,v):(u,v) \in E \wedge {y_u} = {y_v}\} |}}{{|E|}}$ is the fraction of edges which connect nodes that have the same class, i.e., intra-class edges.

The homophily ratio $h$ measures the overall homophily level in the graph and thus we have $h \in [0, 1]$. To be specific, graphs with $h$ closer to 1 tend to have more edges connecting nodes within the same class, or say stronger homophily; on the other hand, graphs with $h$ closer to 0
tend to have more edges connecting nodes in different classes, or say a stronger heterophily.
\section{The Framework}
We first give a brief overview of our approach, and then introduce the proposed new method in specific including detailed descriptions of each component.
\begin{figure*}[ht] 
	\centering
	\includegraphics[width=\linewidth]{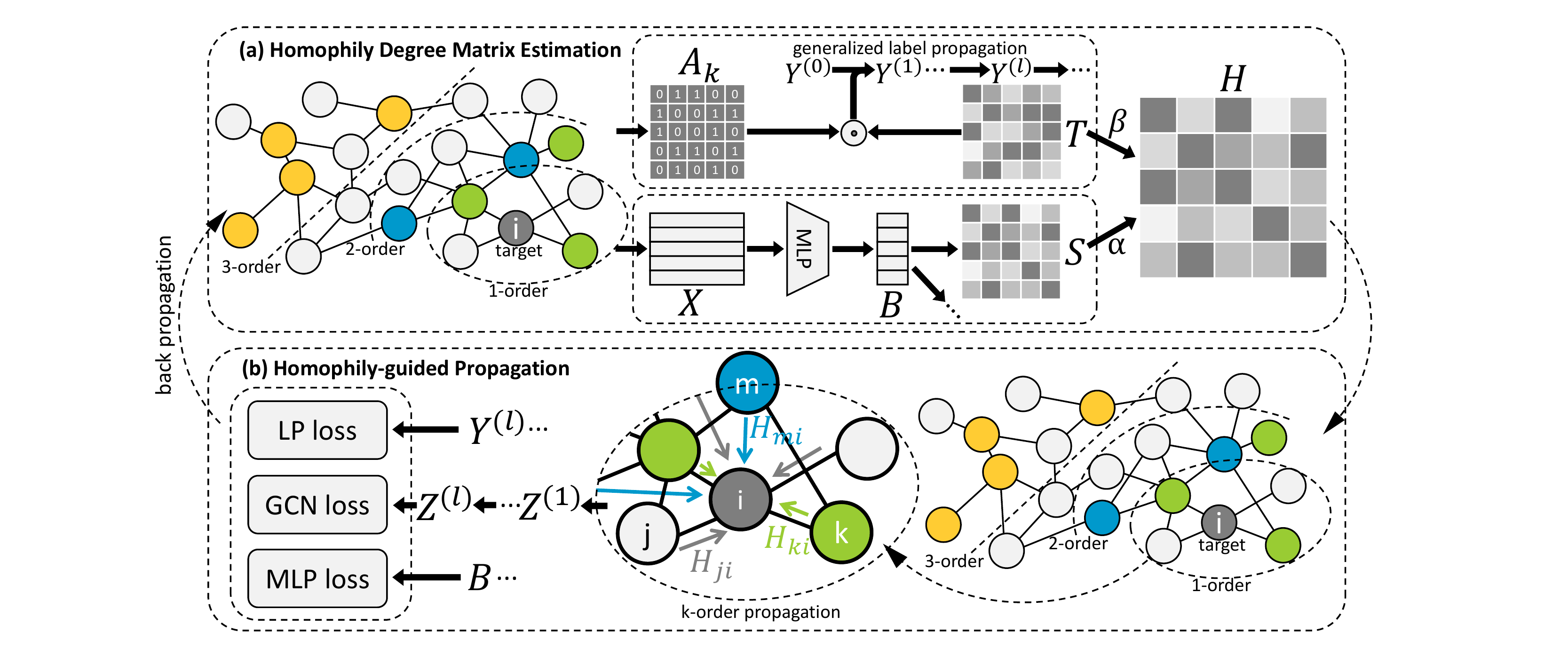}
	\caption{The structure of HOG-GCN, which consists of two components, including (a) homophily degree matrix estimation and (b) homophily-guided propagation.}
	\label{theModel}
\end{figure*}

\subsection{Overview}
To let the propagation mechanism of GCN essentially suitable for both homophily and heterophily, we propose a novel homophily-guided graph convolution framework that can automatically learn the propagation process according to the homophily degree between node pairs, which is called HOG-GCN. In specific, we incorporate a homophily degree matrix into the graph convolution framework for modeling the homophily and heterophily and further use it to adaptively change the propagation process between neighborhoods. The whole structure of the proposed approach is shown in Fig. \ref{theModel}, which consists of two components: homophily degree matrix estimation and homophily-guided propagation. The homophily degree matrix is learned from the attribute and topology information during the propagation process, which is further used to conduct the feature propagation between neighborhoods. In return, the propagation process can help learn better homophily degree matrix through downstream semi-supervised task. Therefore, these two components can be enhanced by each other and trained jointly. For the homophily degree matrix, we extract class-aware information from node attributes and network topology, respectively, and then define the whole homophily degree based on these two types of information. From the perspective of attributes, we leverage multi-layer perceptron to extract class-aware information, since node attributes are not constrained by heterophily. Then, the homophily degree in attribute space can be further calculated. From the perspective of network topology, since it exhibits different degrees of heterophily, we consider making full use of the available label information to capture homophily degree in topology space. To this end, we propose a generalized label propagation technique with a learnable weight matrix, which can reflect homophily degree between node pairs in topology space. The intuition is that the influence between intra-class labels is greater than that between inter-class labels. For the homophily-guided propagation, we introduce the homophily degree matrix into the propagation process which can reveal the underlying distribution of homophily or heterophily in networks, enabling the framework to adaptively change the feature propagation weights according to homophily degree between neighborhoods.
\subsection{Homophily Degree Matrix Estimation}
To adaptively change the propagation process for homophily and heterophily, we learn the homophily degree between node pairs during the propagation process. The homophily degree describes the extent to which two nodes belong to the same class. However, it is difficult to calculate the homophily degree directly from node labels, since only part of label information is available under the semi-supervised task. To this end, we consider estimating the homophily degree between node pairs from attribute space and topology space, respectively, and then combine them with adjustable parameters.

On one hand, from the perspective of attribute space, we apply graph-agnostic multi-layer perceptron (MLP) to extract class-aware information from original node attributes. The $l$-th layer of MLP is defined as:
\begin{equation} \label{2}Z_{m}^{(l)} = \sigma (Z_{m}^{(l - 1)}W_{m}^{(l)})\end{equation}
where $Z_{{m}}^{(0)} = X$, $W_m^{(l)}$ is the learnable weight matrix for MLP, and $\sigma$ the activation function. Denote the output of the final layer of MLP as $Z_m$, then we can obtain the soft assignment matrix ${{B}} \in {\mathbb{R}^{n \times C}}$ as follows :
\begin{equation} \label{3}B = \operatorname{softmax} ({Z_{{m}}})\end{equation}
Each element ${B_{ic}}$ denotes the probability that node $v_i$ belongs to class $c$. Let all parameters of MLP be ${\Theta _{{m}}}$ , then the optimal $\Theta _{{m}}^*$ is obtained by minimizing the following loss of predicted labels by MLP:
\begin{equation} \label{4}\begin{split}
	\Theta _{{m}}^* = \mathop {\operatorname{argmin} }\limits_{{\Theta _{{m}}}} {\mathcal L_{mlp}} 
	= \mathop {\operatorname{argmin} }\limits_{{\Theta _{{m}}}} \frac{1}{{{|V_L|}}}\! \sum\limits_{{v_a} \in {V_L}} \! {J({{\hat b}}_a^{mlp},{y_a})}  \\ 
\end{split}\end{equation}
where ${{\hat b}}_a^{mlp}$  is the predicted labels of $v_a$ by MLP. Since soft assignment matrix $B$ is learned under the guidance of semi-supervised classification, it can capture class-aware information in attribute space. Then based on the matrix $B$ we can calculate the extent that two nodes belong to the same class, called homophily degree matrix, which is defined as:
\begin{equation} \label{5}S = B{B^T}\end{equation}
where ${S_{ij}} = {b_i}b{}_j^T$ denotes the extent to which node $v_i$ and node $v_j$ belong to the same class.

It is worth noting that the homophily degree matrix $S$ is estimated based on original node attributes, which are not constrained by the heterophily of networks. Therefore, it also holds in networks with heterophily or low homophily.

On the other hand, from the perspective of topology space, network topology contains many useful information even if it exhibits strong heterophily. However, it is also difficult to estimate the homophily degree directly based on the topology, as we may not know the distribution of heterophily of network in advance. To address this problem, we further apply label propagation technique to estimate the homophily degree matrix in topology space.

The classic label propagation typically assumes that two connected nodes are more likely to have the same class, and thus propagates labels iteratively between neighborhoods. Let ${Y^{(l)}} = {[y_1^{(l)},y_2^{(l)},...,y_n^{(l)}]^T} \in {\mathbb{R}^{n \times C}}$ denotes the soft label matrix in iteration $l > 0$, where $y_i^{(l)}$ represents predicted label distribution for node $v_i$ in iteration $l$. When $l = 0$, the initial label matrix $Y^{(0)}$ is initialized by the labels of training data, i.e., ${Y^{(0)}} = {[y_1^{(0)},y_2^{(0)},...,y_n^{(0)}]^T}$, which consists of one-hot label indicator vectors $y_i^{(0)}$ for ${{{v}}_{{i}}} \in {V_L}$ or zero vectors otherwise (i.e., unlabeled nodes). Then the label propagation in iteration $l$ is defined as follows: 
\begin{equation} \label{6}\begin{split}
	{Y^{(l)}} = {D^{ - 1}}A{Y^{(l - 1)}}, \hfill \\
	y_i^{(l)} = y_i^{(0)},\forall {v_i} \in {V_L} \hfill \\ 
\end{split} \end{equation}
where $D$ is the degree matrix with entries ${D_{ii}} = \sum\nolimits_j {{A_{ij}}} $. In this equation, all nodes propagate labels to their direct neighborhoods first, and then the labels of all labeled nodes are reset to their initial labels.

However, classic label propagation technique aims to capture the assumption of homophily, which can not adapt to networks with heterophily directly. To capture the homophily degree between node pairs, we generalize classic label propagation with a learnable weight matrix, which is trained under the guidance of labeled data. The key intuition is that the influence between intra-class label is greater than that between inter-class label. Thus the learned weight matrix can be used to represent the extent to which two nodes belong to same class. Since networks exhibit different degrees of heterophily, we perform label propagation over $k$-order structure of network to capture more homophilic nodes (e.g., $k=2$). The $k$-order structure is defined as:
\begin{equation} \label{7}{A_k} = A + {A^2} + ... + {A^k}\end{equation}
Then the generalized label propagation in iteration $l$ is defined as:
\begin{equation} \label{8}{Y^{(l)}} = D_k^{ - 1}({A_k} \odot T){Y^{(l - 1)}}\end{equation}
where $D_k^{ - 1}$ is the diagonal degree matrix for matrix ${A_k} \odot T$. In the equation above, all nodes propagate labels to their $k$-order neighborhoods according the learnable edge weights. Then we can learn the optimal edge weights $T^*$ by minimizing the loss of predicted labels by generalized label propagation:
\begin{equation} \label{9}
	{T^{{*}}}{{ = }}\mathop {\operatorname{argmin} }\limits_T { \mathcal L_{lp}} \\ 
	= \mathop {\operatorname{argmin} }\limits_T \frac{1}{{{|V_L|}}}\sum\limits_{{v_a} \in {V_L}} {J(\hat y_a^{lp},{y_a})}  \\ 
\end{equation}
where $J$ is the cross-entropy loss, $\hat y_a^{lp}$ and ${y_a}$ are the predicted label distribution of $v_a$ by generalized label propagation and the true one-hot label of $v_a$, respectively. Note that we do not add to or remove edges from graphs, but only learn the weights of existing $k$-order neighborhoods. The optimal $T^*$ maximizes the probability that each node is correctly classified by generalized label propagation, thus also increases the intra-class influence. This reflects the extent to which two nodes belongs to the same class. In order to form an end-to-end pattern, here we take the weight matrix $T$ as the homophily degree matrix estimated from topology space.

At last, to make the model more effective and robust, we combine the hompily degree matrix estimated from attribute space and topology space with adjustable parameters as follows:
\begin{equation} 
	\label{10}H = \alpha S + \beta T
\end{equation}
where $\alpha$ and $\beta$ are hyper-parameters. Also of note, matrix $S$ contains homophily degree between any node pairs while $T$ learns homophily degree within k-order neighborhoods. It does not influence the performance, as we will filter the entries that do not involve in propagation process.
\subsection{Homophily-guided Propagation}
The  \textsl{core} of our approach is adaptively learning the propagation process for homophily and heterophily. The homophily degree matrix which represents the extent to which two nodes belong to the same class can reflect the underlying distribution of homophily or heterophily of networks. Therefore, we incorporate the learnable homophily degree matrix into graph convolution framework to automatically change the propagation weights between neighborhoods according to homophily degree. To be specific, during the propagation process, we aim to increase the feature influence between underlying intra-class nodes and reducing feature influence between underlying inter-class nodes according to the homophily degree matrix. To achieve this, we assign different weights to different neighborhoods according to the homophily degree, which can distinguish the homophily or heterophily between neighborhoods. Similarly, as we may not know the underlying distribution of heterophily in advance, we perform feature propagation over the $k$-order neighborhoods, to capture more homophilic nodes. In addition, we deal with ego-representation and neighborhood-representation separately to preserve more personalized information. Then, the feature propagation process of the proposed method HOG-GCN in iteration $l$ is given by:
\begin{equation}
	\label{11}{Z^{(l)}} = \sigma (\mu {Z^{(l - 1)}}W_e^{(l)} + \xi {\hat D^{ - 1}}{A_k} \odot HZ^{(l - 1)}W_n^{(l)})
\end{equation}
where $\mu$ and $\xi$ denote the weights of ego-representation and neighborhood-representation respectively, $\hat D$ is the diagonal degree matrix, ${Z^{(0)}} = X$ is the original node attributes. $\sigma$ is the activation function.

Also of note, our proposed graph convolution framework can be taken as learning proper attention weights. The most significant difference between our approach and those feature-based attention methods is that, the attention weights are learned based on feature similarity alone, while our proposed method measures the edge weights according to the underlying homophily degree, which is more task-oriented. In addition, it is often believed that the 2-hop neighborhoods of a node $v$ is always homophily-dominant in expectation \cite{H2GCN}. So, we also set $k = 2$ in this work since it yields best performance in experiments and has relatively lower complexity meanwhile.
\subsection{Optimization Objective}
The whole framework consists of two components: homophily degree matrix estimation and homophily-guided propagation. The first component contains MLP and the generalized label propagation. The objective functions of MLP and generalized label propagation are give by Eq. (\ref{4}) and Eq. (\ref{9}). In the second component, we incorporate the homophily degree matrix into the graph convolution framework. Denote all parameters of graph convolution operation as ${\Theta _{{g}}}$, then we can get the optimal $\Theta _{{g}}^*$ according the final output $Z$ of HOG-GCN:
\begin{equation} \label{12}\begin{split}
	&{R} = \operatorname{softmax}(Z) \\ 
	\Theta _g^* = \operatorname{argmin} &{\mathcal L_{gcn}} 
	 = \mathop {\operatorname{argmin}} \limits_{{\Theta _g}} \frac{1}{{|V_L|}} \! \sum\limits_{{v_a} \in {V_L}} \! {J(r_a^{gcn},{y_a})}  \\ 
\end{split} 
\end{equation}
In this model, the homophily degree matrix is learned from attribute and topology information during the propagation process and further used to conduct feature propagation. In return, the propagation process can help learn better homophily degree matrix. That is, these two components are enhanced by each other. Therefore, we combine these objectives to train the whole process in an end-to-end fashion as:
\begin{equation} \label{13}\Theta _g^*,\Theta _{{m}}^*,{T^*} = \mathop {\operatorname{argmin}\;}\limits_{{\Theta _g},{\Theta _m},T} {\mathcal L_{gcn}} + \lambda {\mathcal L_{mlp}} + \gamma {\mathcal L_{lp}}
\end{equation}
where $\lambda$, $\gamma$ are balance hyper-parameters. In this way, the feature propagation process is guided by current homophily degree matrix, and the result of propagation can conduct the learning of homophily degree matrix through semi-supervised classification.
\subsection{Theoretical Analysis}
In this section we prove that our new approach can constrain the similarity of representations between nodes within k-order neighborhoods according to homophily degree between them. In other words, nodes with higher homophily degree have more similar representation (Theorem 1).

\noindent \textbf{Theorem 1.} Denote the output of HOG-GCN as $Z$, then its propagation process can be taken as the minimization of the objective:
\begin{equation} \label{14}{{O}} = \frac{1}{2}\sum\limits_{{v_i} \in V}\!{\sum\limits_{{v_j} \in {N_k}({v_i})} {{{{H}}_{ij}}{{\left\|{Z_i} - {Z_j}\right\|}^2}} } 
\end{equation}
where $N_k(v_i)$ denotes the $k$-order neighborhoods of node $v_i$, $Z_i$ and $Z_j$ are representations for node $v_i$ and $v_j$ respectively.\\
\textbf{Proof:} We can rewrite the above equation as:
\begin{equation} \label{15}\begin{split}
	O =& \frac{1}{2}\sum\limits_{{v_i} \in V} {\sum\limits_{{v_j} \in {N_k}({v_i})} {{{H}_{ij}}{{\left\|{Z_i} - {Z_j}\right\|}^2}} }  \\ 
	=& \operatorname{tr}({Z^T}(\hat D - {A_k} \odot H)Z) \\ 
\end{split} 
\end{equation}
where ${{\hat D}}$ is the degree matrix for ${A_k} \odot H$. By setting derivative of Eq. (\ref{15}) with respect to $Z$ to zero (assume $H$ is a symmetric matrix), we have:
\begin{equation} \label{16}\begin{split}
	\frac{{\partial {O}}}{{\partial {Z}}} =& 2(\hat D - {A_k} \odot H)Z = 0 \\ 
	\Downarrow&  \\ 
	Z =& {{\hat D}^{ - 1}}({A_k} \odot H)Z \\ 
\end{split} 
\end{equation}
Then, it can be explained as a limit distribution where
\begin{equation} \label{17}{Z_{\operatorname{limit}}} = {\hat D^{ - 1}}({A_k} \odot H){Z_{\operatorname{limit}}}
\end{equation}
We use the following iterative form to approximate the limit with $l \to \infty$:
\begin{equation} \label{18}{Z^{(l)}} = {\hat D^{ - 1}}({A_k} \odot H){Z^{(l - 1)}}
\end{equation}
When ignoring the non-linear transformation and initialize $Z^{(0)}=XW$, we have:
\begin{equation} \label{19}\begin{split}
	{Z^{(l)}} =& {{\hat D}^{ - 1}}({A_k} \odot H){Z^{(l - 1)}} \\ 
	 =& [{{\hat D}^{ - 1}}({A_k} \odot H){]^2}{Z^{(l - 2)}} \\ 
	=& {{[}{{\hat D}^{ - 1}}({A_k} \odot H)]^l}{Z^{(0)}} \\ 
	=& {{[}{{\hat D}^{ - 1}}({A_k} \odot H)]^l}XW \\ 
\end{split}\end{equation}
which corresponds to the propagation process of HOG-GCN.


\section{Experiments}
We first give the experimental setup, then compare our model HOG-GCN with the state-of-the-art methods on transductive node classification and visualization. Last, we give the parameter analysis and homophily degree matrix analysis.

\subsection{Experimental setup}
\textbf{Datasets.} We evaluate the performance of the proposed HOG-GCN and existing methods on seven real-world datasets. To demonstrate that HOG-GCN can adaptively learn propagation mechanism for both homophily and heterophily, we use four heterophilic networks and three homophilic networks for evaluation. The heterophilic networks include three web page datasets, Cornell, Texas and Wisconsin \cite{Geom-GCN}, and a film industry dataset Film \cite{DatasetFilm}. The homophilic networks include Cora, Citeseer and Pubmed which are all citation networks \cite{DatasetCite1, DatasetCite2}. The detailed statistics of these datasets is summarized in Table \ref{datasets}, where H.R. represents the homophily ratio of networks.\\
			\begin{table}[h]
	\footnotesize
	\centering
	\resizebox{\columnwidth}{!}{
		\begin{tabular}{lccccccc}
			
			\toprule
			\multicolumn{1}{l}{\textbf{Datasets}} & \textbf{Texa.} & \textbf{Wisc.} & \textbf{Corn.} & \textbf{Film} & \textbf{Cora} & \textbf{Cite.} & \textbf{Pubm.} \\
			\midrule     
			Nodes            & 183            & 251                & 183              & 7600          & 2708          & 3327              & 19717           \\
		Edges            & 309            & 499                & 295              & 
		33544         & 5429          & 4732              & 44338           \\
		Features         & 1703           & 1703               & 1703             & 931           & 1433          & 3703              & 500             \\
		Classes          & 5              & 5                  & 5                & 5             & 7             & 6                 & 3               \\
		H.R.  & 0.09           & 0.19               & 0.30             & 0.22          & 0.81          & 0.74              & 0.80 \\ 
			\bottomrule
		\end{tabular}	
	}
	\caption{Statistics of datasets}
	\label{datasets}
\end{table}
	
\noindent \textbf{Baselines.} We compare our proposed approach HOG-GCN with the following baselines: (1) MLP, which only uses attribute information; (2) DeepWalk \cite{DeepWalk}, which uses network topology  information alone through random walks; (3) traditional GNN models : GCN \cite{GCN} and GAT \cite{GAT}, which work under the assumption of homophily; and (4) GNN models tackling heterophily: Geom-GCN \cite{Geom-GCN}, H2GCN \cite{H2GCN}, CPGNN \cite{CPGNN}, GPR-GNN \cite{GPR-GNN} and AM-GCN \cite{AM-GCN}. Particularly, as CPGNN has four different variants, we choose the best two for comparison.

\begin{table*}[h]
	\centering
	\scriptsize
	\begin{tabular}{P{2.2cm}|*{7}{P{1.1cm}}}
		\specialrule{1pt}{0pt}{0pt}
		\textbf{\tiny{Datasets / Accuracy (\%)}} & \textbf{Texas}      & \textbf{Wisconsin}  & \textbf{Cornell}    & \textbf{Film}       & \textbf{Cora}       & \textbf{Citeseer}   & \textbf{Pubmed}\\
		GCN              & 54.05$\pm$4.36          & 50.39$\pm$7.55          & 53.78$\pm$8.59          & 28.78$\pm$1.48          & 86.48$\pm$1.43          & 73.58$\pm$1.37          & 87.34$\pm$0.65\\
		MLP              & 81.08$\pm$4.83          & \underline{85.49$\pm$3.53}    & \underline{83.24$\pm$7.03}    & 36.58$\pm$1.44          & 71.29$\pm$1.60          & 66.96$\pm$2.61          & 86.48$\pm$0.63\\
		GAT              & 57.30$\pm$3.38          & 54.31$\pm$5.62          & 54.59$\pm$7.33          & 28.99$\pm$1.44          & \underline{87.16$\pm$1.17}    & 75.64$\pm$1.95          & 85.25$\pm$0.60\\
		DeepWalk         & 49.19$\pm$3.38          & 53.51$\pm$5.10          & 44.12$\pm$9.12          & 23.74$\pm$0.56          & 80.08$\pm$1.84          & 53.59$\pm$2.63          & 81.14$\pm$0.54\\
		H2GCN            & 79.73$\pm$7.27          & 82.55$\pm$4.33          & 78.38$\pm$4.35          & 36.71$\pm$1.41          & 86.48$\pm$1.63          & 75.56$\pm$2.18          & 88.77$\pm$0.65\\
		CPGNN-MLP        & 79.73$\pm$6.54          & 84.53$\pm$6.48          & 73.51$\pm$6.02          & \underline{36.73$\pm$1.03}    & 84.57$\pm$1.59          & 72.10$\pm$2.70          & 87.67$\pm$0.72\\
		CPGNN-Cheby      & 74.32$\pm$7.38          & 81.76$\pm$6.74          & 63.51$\pm$5.83          & 35.51$\pm$1.85          & \textbf{87.18$\pm$1.13} & 75.52$\pm$1.84          & \textbf{89.08$\pm$0.67}\\
		GPR-GNN          & \underline{84.59$\pm$4.37}    & 83.92$\pm$3.14          & 82.97$\pm$5.68          & 36.47$\pm$1.38          & 86.70$\pm$1.03          & 75.12$\pm$1.98          & 87.38$\pm$0.63\\
		AM-GCN           & 78.38$\pm$7.25          & 81.76$\pm$4.96          & 78.38$\pm$4.98          & 33.60$\pm$1.17          & 86.66$\pm$1.36          & \underline{76.01$\pm$1.90}    & 86.78$\pm$0.60\\
		Geom-GCN         & 66.22$\pm$6.65          & 62.55$\pm$5.22          & 55.68$\pm$8.04          & 32.39$\pm$1.49          & 84.91$\pm$1.12          & 73.16$\pm$1.92          & 88.41$\pm$0.63\\
		\specialrule{0.5pt}{1pt}{1pt}
		HOG-GCN          & \textbf{85.17$\pm$4.40} & \textbf{86.67$\pm$3.36} & \textbf{84.32$\pm$4.32} & \textbf{36.82$\pm$0.84} & 87.04$\pm$1.10          & \textbf{76.15$\pm$1.88} & \underline{88.79$\pm$0.40}\\ 
		\specialrule{1pt}{0pt}{0pt}     
	\end{tabular}
	
	\caption{Classification results with mean value and standard deviation. The best result is bold and the second best is underlined.}
	\label{classification}
\end{table*}

\begin{figure*}[ht]
	\centering
	
	\subfigure[GCN] {  
		\includegraphics[width=0.33\columnwidth]{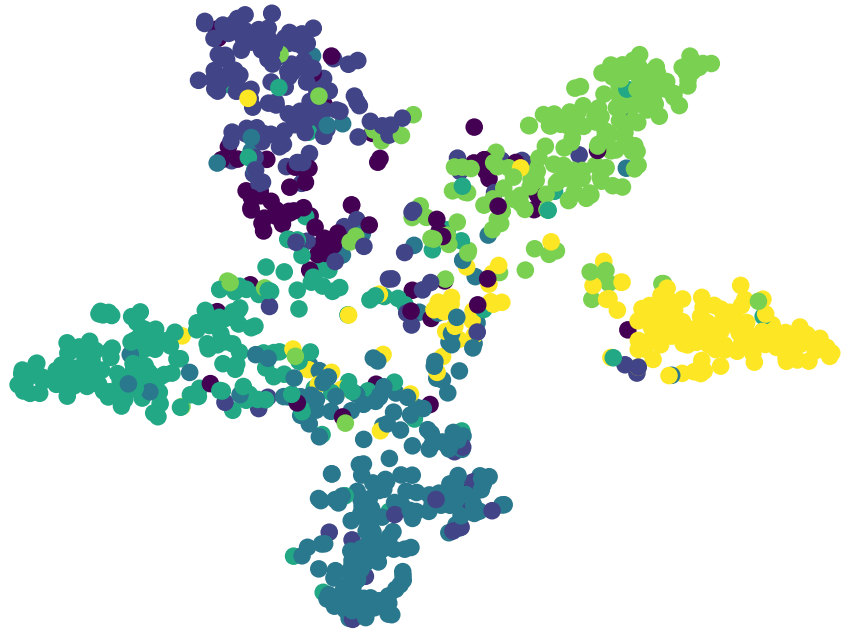}
	}
	\subfigure[H2GCN] {
		\includegraphics[width=0.33\columnwidth]{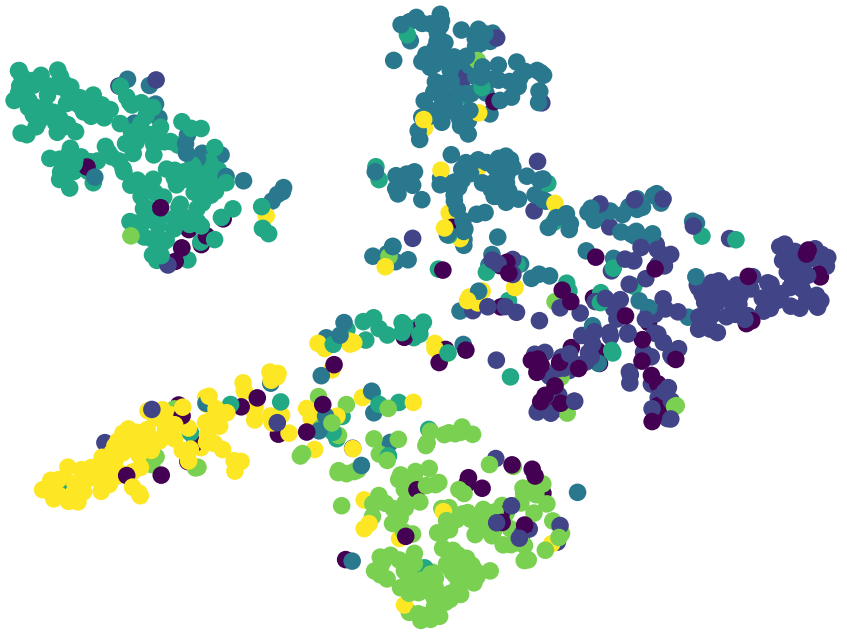}
	}
	\label{attcc}
	\subfigure[CPGNN] {
		\includegraphics[width=0.33\columnwidth]{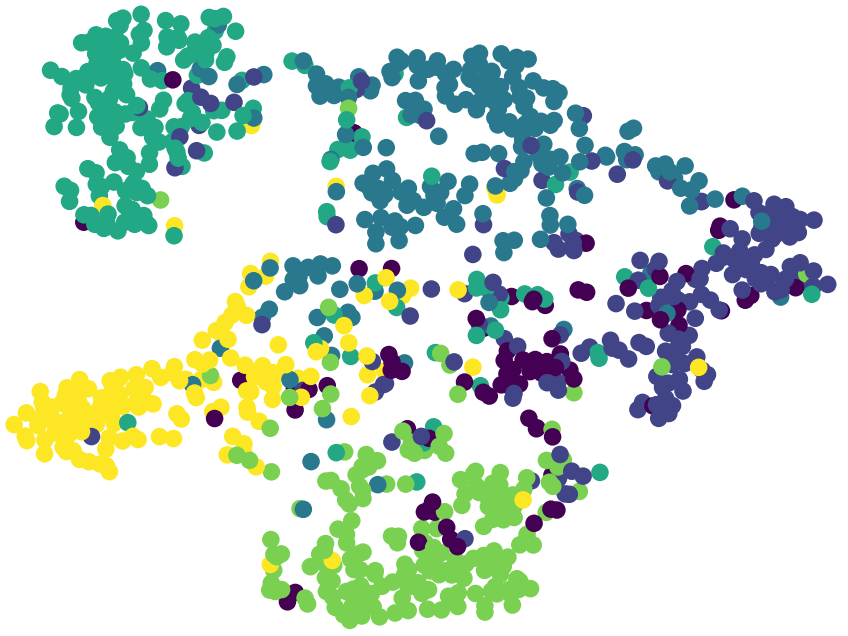}
	}
	\label{attbb}
	\subfigure[GPRGNN] {
		\includegraphics[width=0.33\columnwidth]{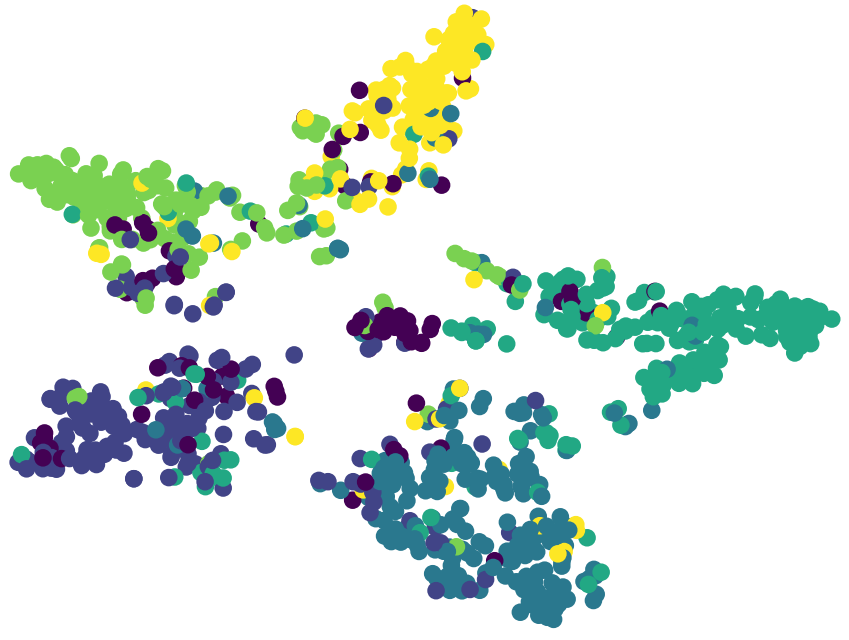}
		
	}
	\subfigure[HOG-GCN] {
		\includegraphics[width=0.33\columnwidth]{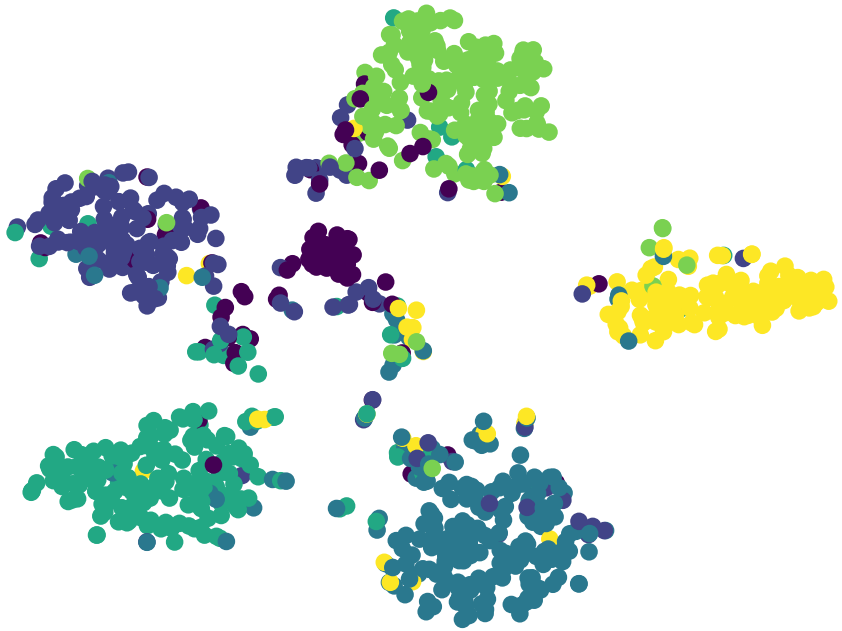}
		
	}
	\caption{Visualization results on Citeseer dataset}\label{visualization}
\end{figure*}

\noindent \textbf{Parameter Setup.} 
For all datasets, we generate 10 random splits for training, validation and test. For each split we select 48\% of nodes in each class to form the training set, 32\% of nodes for the validation set and the remaining as the test set. For a fair comparison, all methods share the same 10 random splits. All the parameters of the baseline methods were set as what were used by their authors. 
In our approach HOG-GCN, for the homophily degree matrix estimation, we use a two-layer MLP with 512 units in the hidden layer. For the homophily-guided propagation, we use two-layer graph convolution operation with 256 units in the hidden layer. We set $\alpha$ to 1 and $\beta$ to 0.1, and set both $\gamma$ and $\mu$ to 1. We adopt Adam optimizer \cite{adam} and the default initialization in pytorch.

\subsection{Node Classification}
The node classification results are reported in Table \ref{classification}. We use mean accuracy as the evaluation metric along with the standard deviation of 10 splits. We have the following observations.
\begin{itemize}
\item Our approach HOG-GCN outperforms all the other methods compared on all of the four heterophilic networks, i.e., Texas, Cornell, Wisconsin and Film. It demonstrates the importance of incorporating and learning the homophily degree matrix into graph convolution framework for automatically changing the propagation process. To be specific, HOG-GCN significantly outperforms the traditional GNN models, i.e., GCN and GAT, by 26.49\% and 24.47\% on average, since they cannot generalize to the scenario of heterophily (and are even defeated by MLP which only uses attributes). Compared with methods that focus on tackling heterophily, such as H2GCN, Geom-GCN, CPGNN and GPR-GNN, our HOG-GCN also achieves an improvement between 3.9\% and 19.03\% in terms of mean accuracy. These results demonstrate the effectiveness of HOG-GCN in the scenario of heterophily.   
\item On homophilic networks (i.e., Cora, Citeseer, Pubmed), the proposed HOG-GCN performs better or comparable to the baselines. To be specific, HOG-GCN performs best on Citeseer, and achieves the second best on Pubmed. Notably, HOG-GCN outperforms GCN and GAT which have an implicit assumption of strong homophily on these three homophilic networks by 1.52\% and 0.98\% on average. These results demonstrate that our method has the best performance in the scenario of heterophily while maintaining comparable or better performance in the scenario of homophily, and further validate the effectiveness and robustness of the proposed approach.
\end{itemize}
\begin{figure}[ht]
	\centering
	\subfigure[Heterophilic networks] {
		\centering 
		\includegraphics[width=0.41\columnwidth]{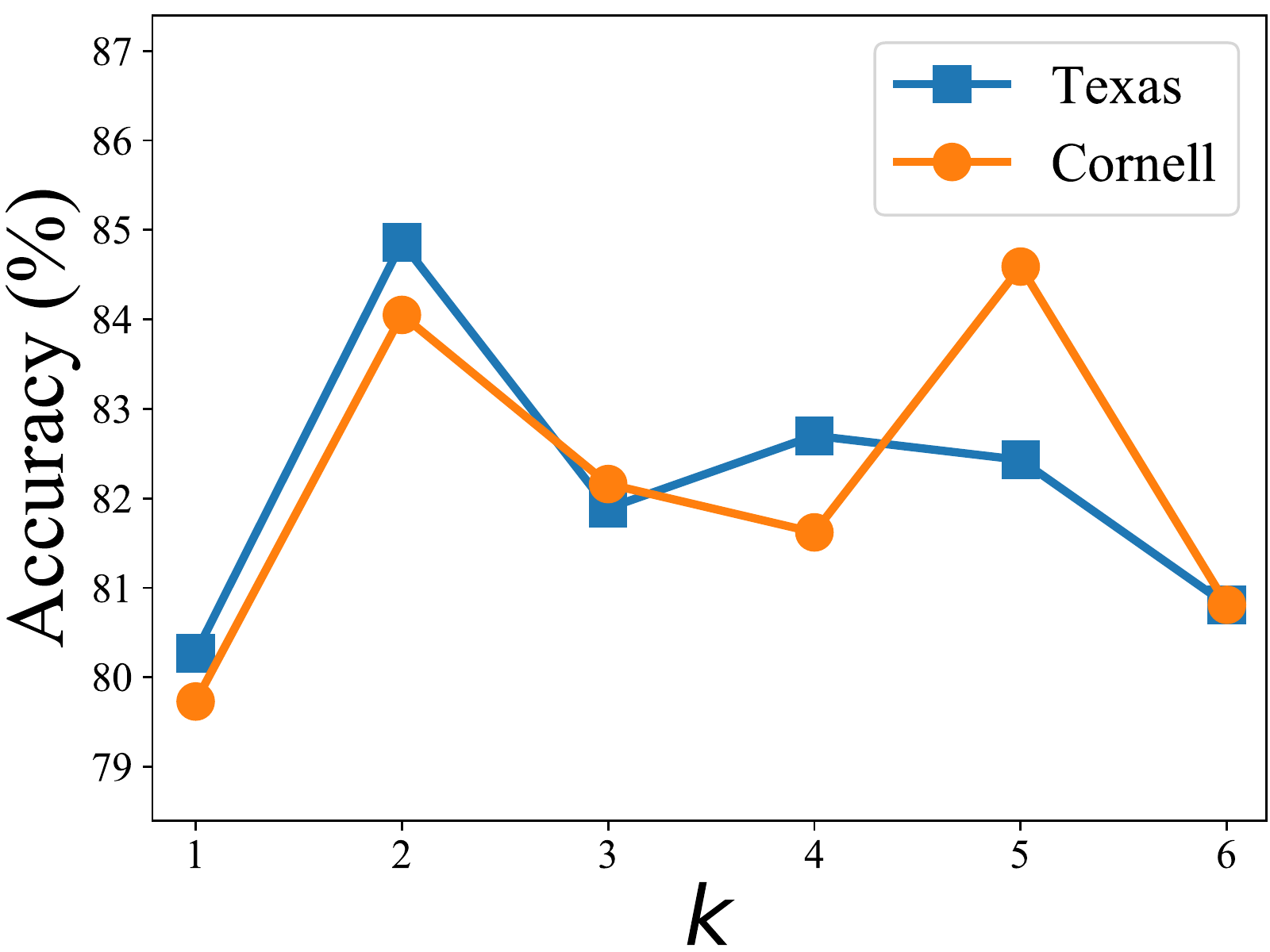}
		\label{feature_acc}
	}
	\subfigure[Homophilic networks] {
		\centering
		\includegraphics[width=0.41\columnwidth]{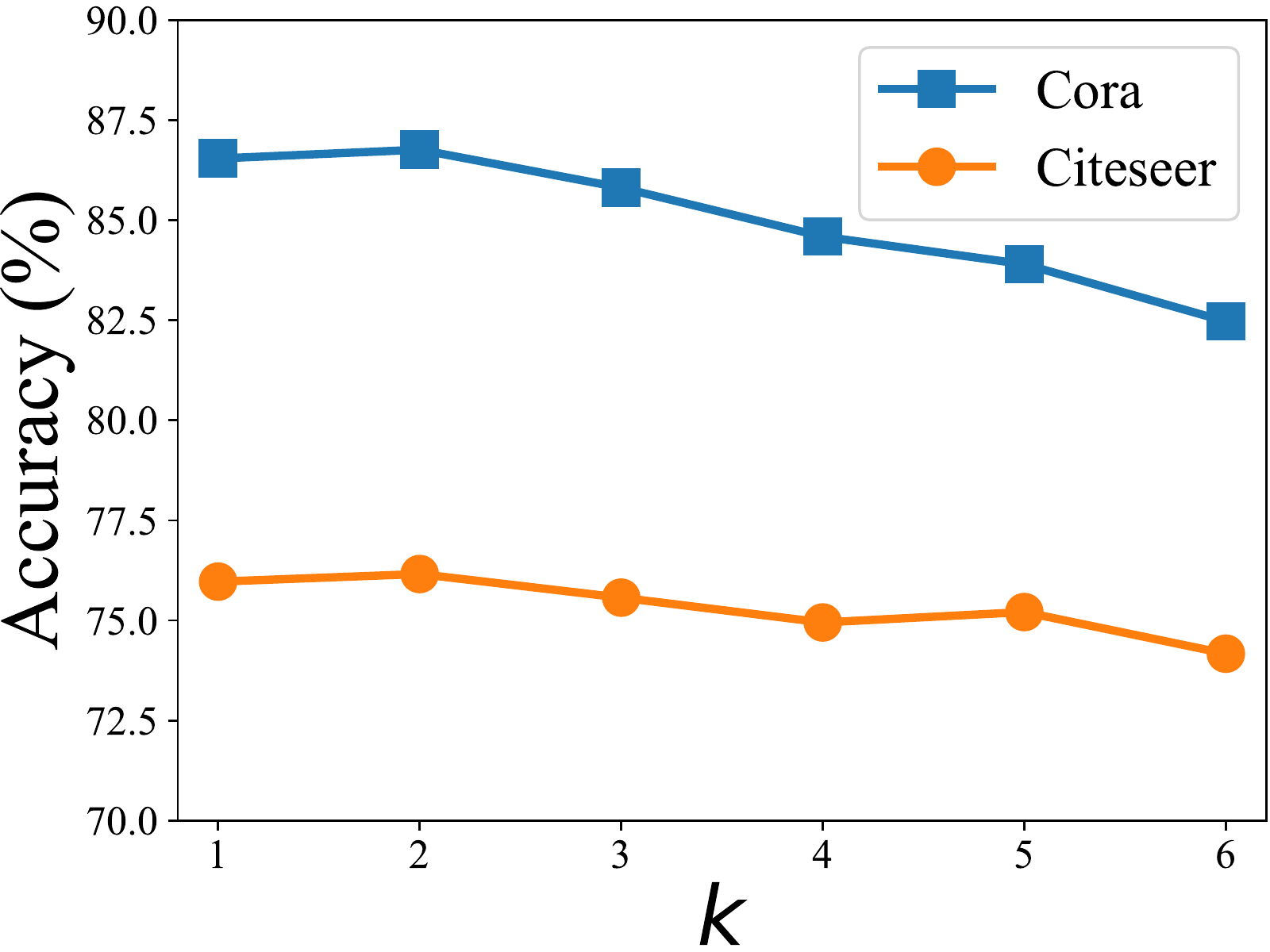}
		\label{feature_acc}
	}
	\caption{Analysis results of order $k$}\label{k_analysis}
\end{figure}
\subsection{Visualization}
For intuitively illustrating that our approach can gain better results, we use t-SNE \cite{t-SNE}, which can project the learned node representations onto a two-dimensional space, to visualize the derived representations. Fig. \ref{visualization} shows the visualization results of  GCN, H2GCN, CPGNN, GPR-GNN and our proposed HOG-GCN on Citeseer dataset as an illustrative example, where nodes with same color have same class label. As shown, the visualization results of GCN and CPGNN are less satisfactory in this case, since points with same color are dispersed and some points with different color are mixed with each other. The results of GPR-GNN and H2GCN are relatively better but the borders between different classes are still not so clear. Obviously, the visualization of our HOG-GCN performs much better, where the representations have the highest intra-class similarity and form more discernible clusters. The results of visualization validate that our proposed method can gain better result, and demonstrate the effectiveness of our theoretical analysis.

\begin{figure}[t]
	\centering
	\subfigure[Texas] {
		\centering
		\includegraphics[width=0.45\columnwidth]{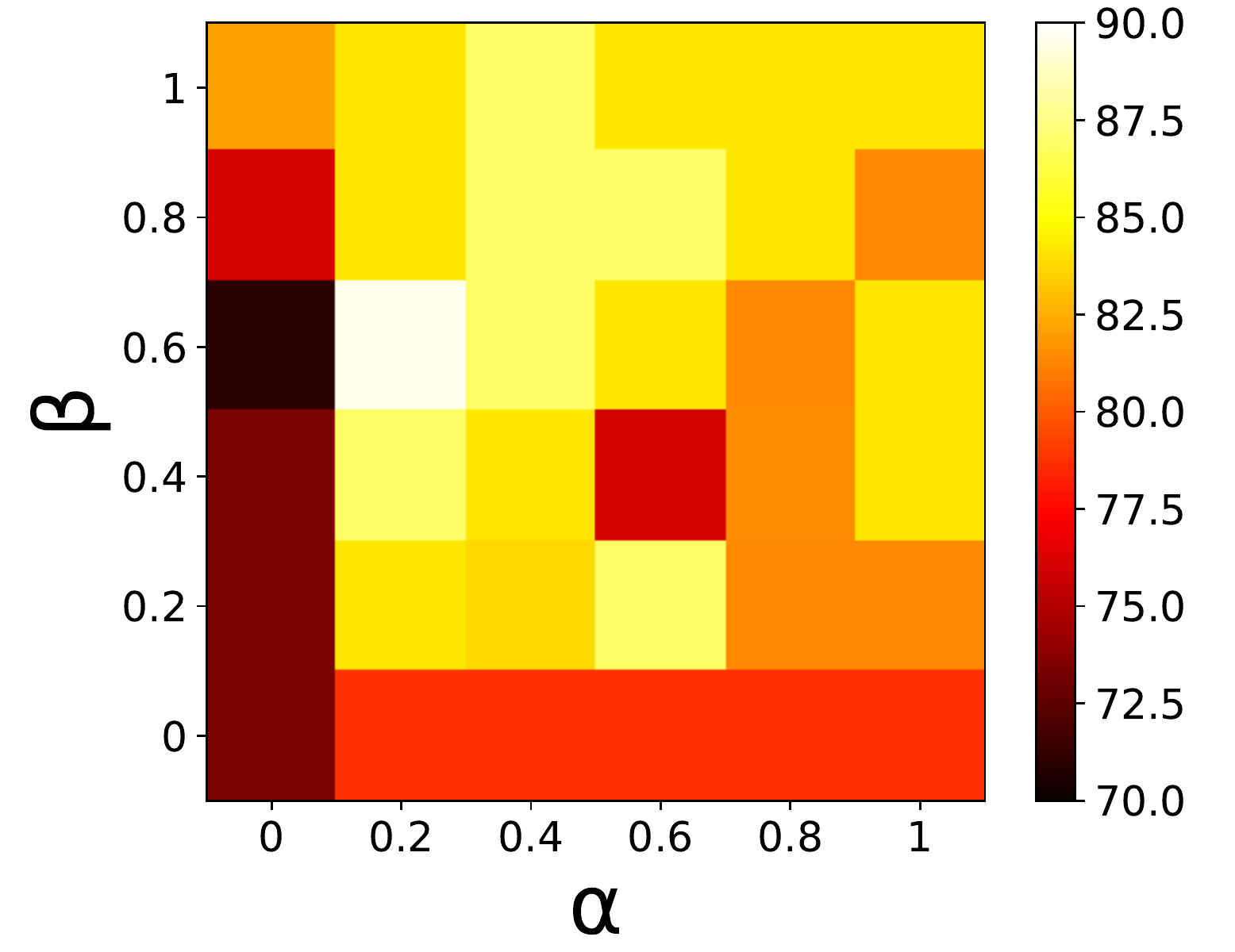}
	}
	\subfigure[Cora] {
		\centering
		\includegraphics[width=0.45\columnwidth]{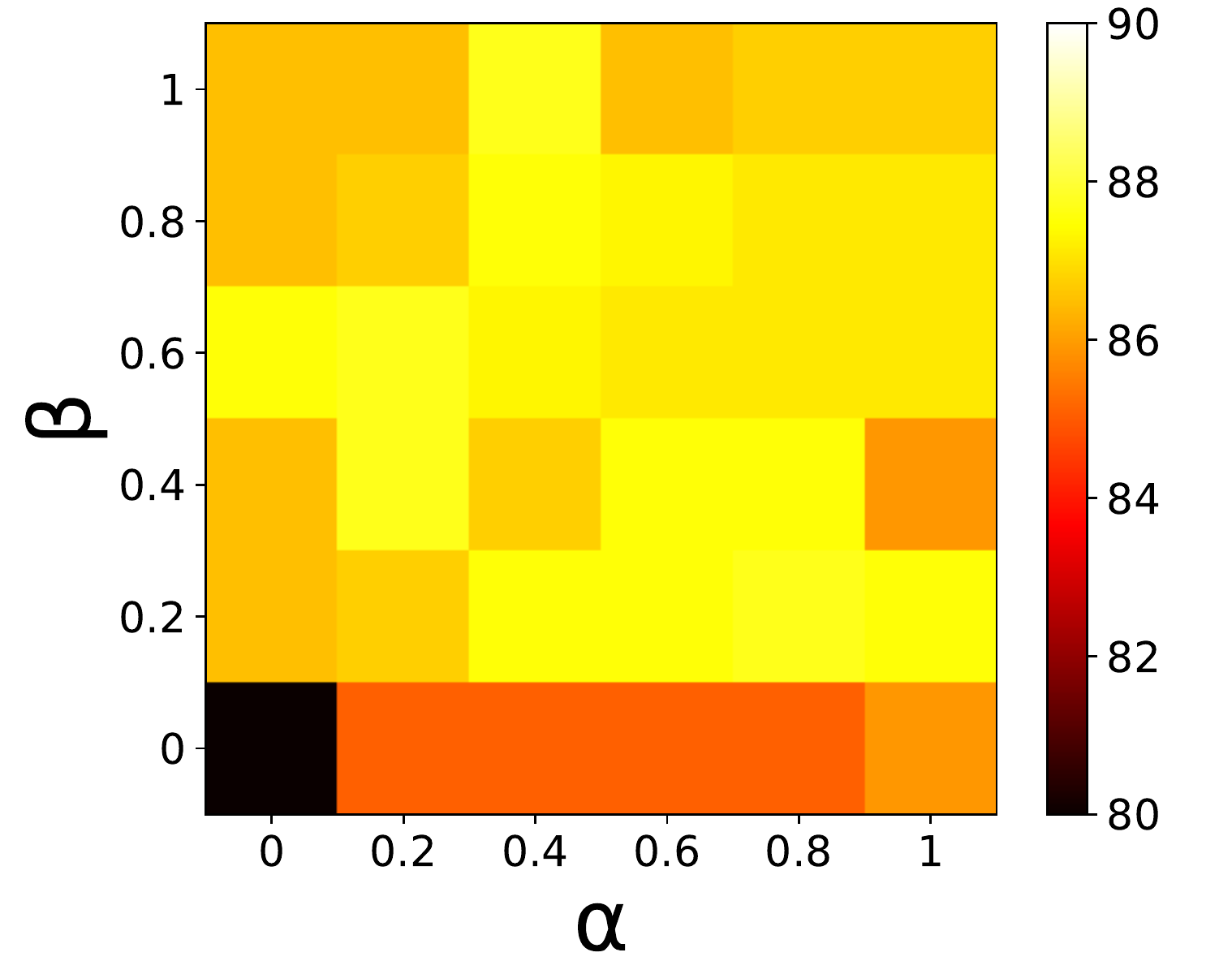}
	}
	\caption{Analysis results of weight $\alpha$ and $\beta$}\label{degree_weights}
\end{figure}
\subsection{Parameter Analysis}
We also investigate the sensitivity of parameters in HOG-GCN. We take Texas, Cornell, Cora, Citeseer as examples, where Texas and Cornell exhibit heterophily while Cora and Citeseer exhibit homophily.

\noindent \textbf{Analysis of order $k$.} In our proposed method, we conduct generalized label propagation and homophily-guided feature propagation over the network topology within k-order neighborhoods, since we may not know the distribution of heterophily of network. Here, we vary the order $k$ from 1 to 6 and report the classification results. The results are shown in Fig. \ref{k_analysis}. As shown, results on Texas and Cornell have a significant improvement when $k = 2$ than $k =1$. While on Cora and Citeseer, the method only has a slight improvement when $k = 2$ than $k =1$, and the accuracy drops as increasing $k$ from 2 to 6. Based on these results, we conclude that on heterophilic networks where most directly connected nodes are from different classes, we often need a proper receptive field to capture enough intra-class information; while on homophilic networks where most connected nodes are from the same class, direct neighborhoods may be enough to provide intra-class information. For both homophilic and heterophilic networks, aggregating much information from higher-order neighborhoods will mix much noise and irrelated information that will negatively influence the prediction of nodes.

\noindent \textbf{Analysis of weights $\alpha$ and $\beta$.}
In our model, $\alpha$ and $\beta$ represent the weight of homophily degree matrix estimated from attribute space and topology, respectively. Here we investigate the influence of $\alpha$ and $\beta$ on node classification by varying them from 0 to 1. The results are shown in Fig. \ref{degree_weights}. On Texas dataset, the performance is relatively poor when ignoring attribute information or topology information (i.e., $\alpha$ is 0 or $\beta$ is 0). The method performs best when $\alpha$ is 0.4 and $\beta$ is 0.6, demonstrating that combining attribute information and topology information can learn better homophily degree matrix and further obtain better performance. On Cora dataset, the performance is relatively stable, demonstrating that homophilic networks may not need fine-grained guidance during the propagation process. The results demonstrate the importance of combining node attributes and network topology in the estimation of homophily degree matrix, and further validate the effectiveness and robustness of the proposed approach. 

\begin{figure}[t]
	\centering
	\subfigure[Texas] {
		\centering
		\includegraphics[width=0.41\columnwidth]{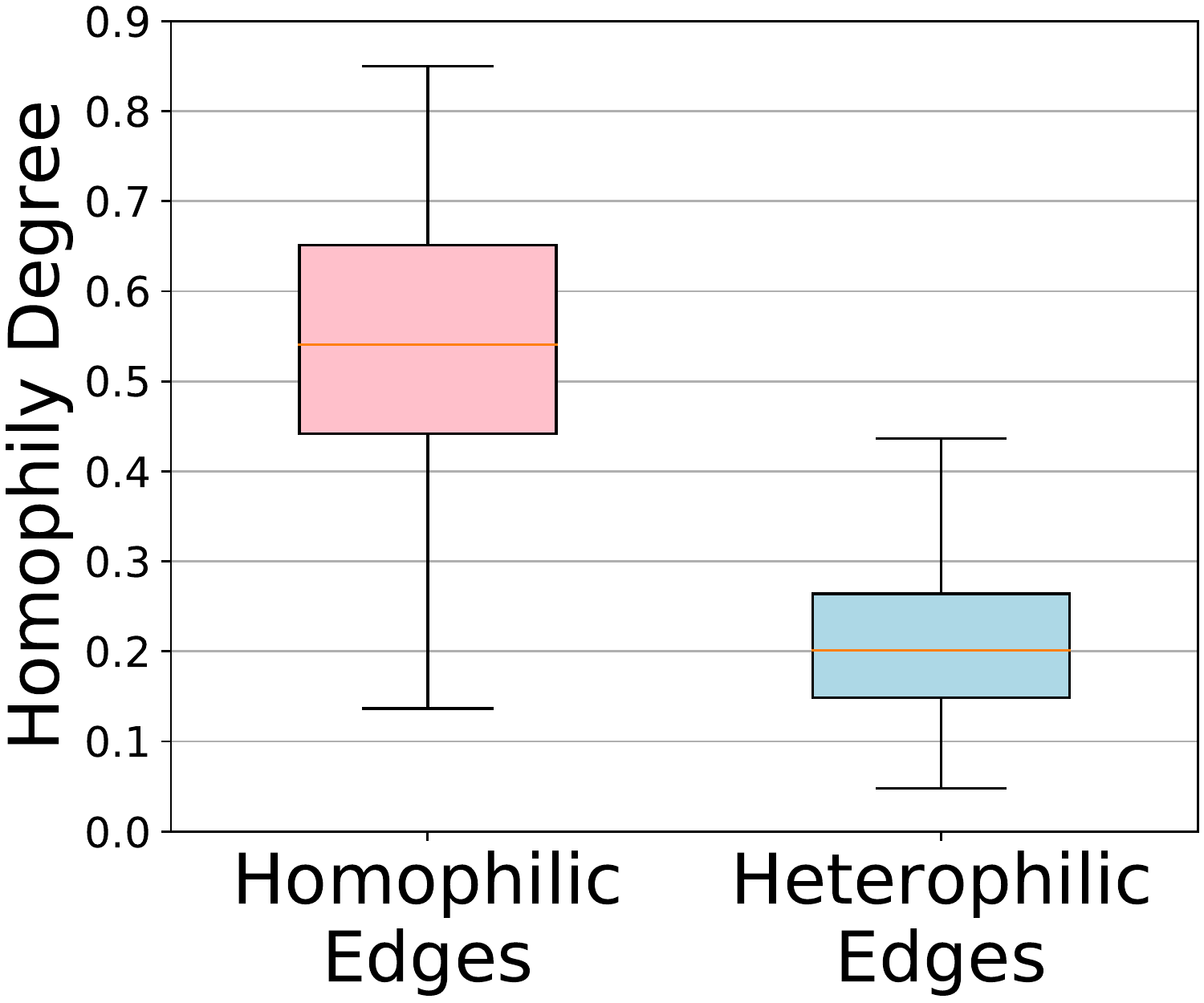}
	}
	\subfigure[Cora] {
		\centering
		\includegraphics[width=0.41\columnwidth]{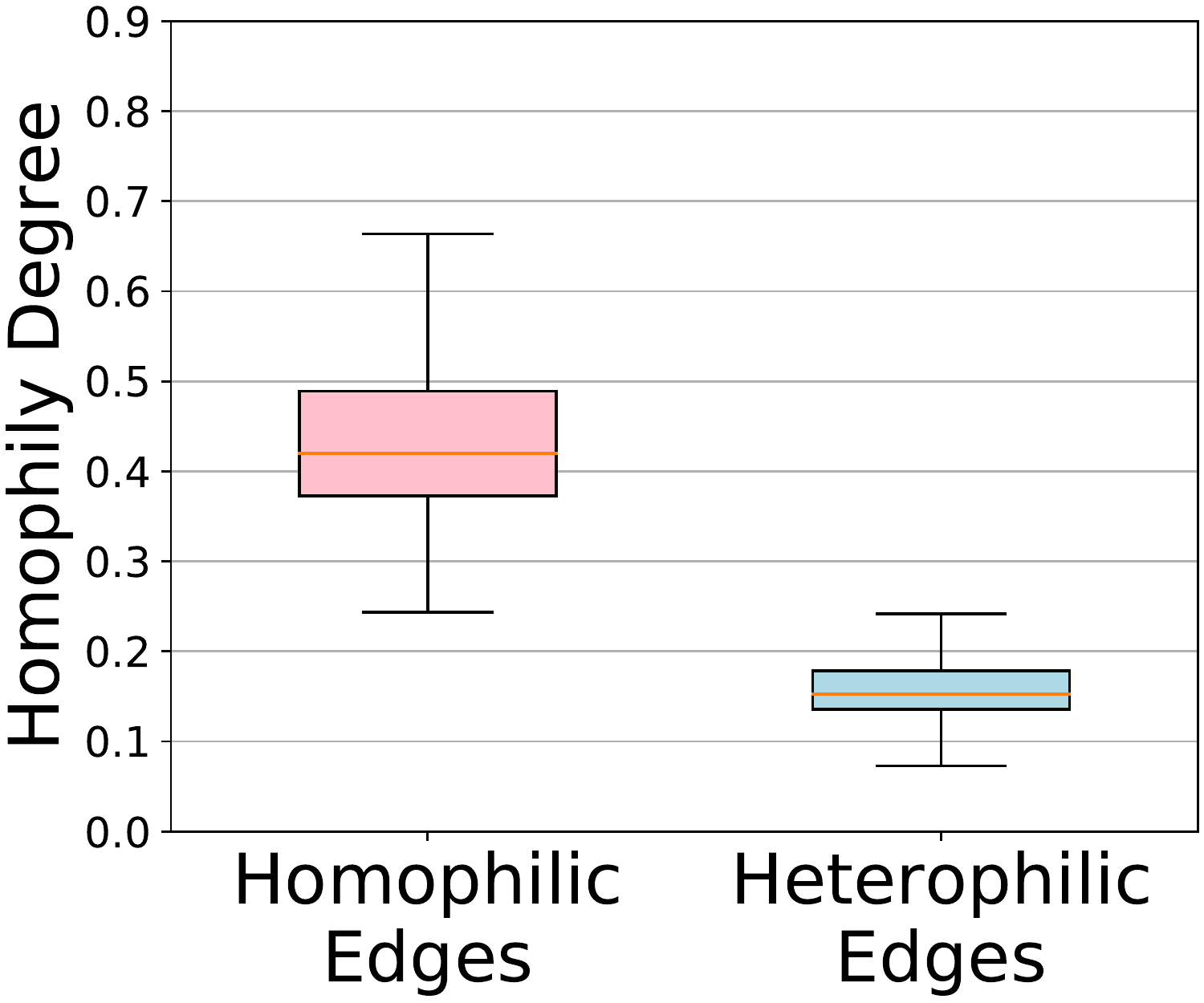}
	}
	\caption{Analysis results of homophily degree distribution}\label{distribution_analysis}
\end{figure}
\subsection{Homophily Degree Matrix Analysis}
In the proposed graph convolution framework, we incorporate a learnable homophily degree matrix, which can reflect the homophily or heterophily of networks, to adaptively change the propagation process. Therefore, in order to investigate whether the homophily degree matrix learned by our proposed model are meaningful, here we analyze the homophily degree distribution on homophilic edges (i.e., two nodes with same class) and heterophilic edges (i.e., two nodes with different classes) within 2-order neighborhoods, respectively. Fig. \ref{distribution_analysis} shows the analysis results on a heterophilic network Texas and a homophilc network Cora.

As shown, for both homophilc and heterophilic networks, our approach can adaptively learn higher homophily degree between node with the same class, and lower homophily degree between nodes with different classes. During the feature propagation process, the homophilic edges are given greater weights while the heterophilic edges are given smaller weights according to homophily degree matrix. The results demonstrate that our approach can adaptively learn different propagation process for both homophily and heterophily, enabling it to go beyond the strong assumption of homophily. 
\section{Conclusion}
In this paper, we propose a novel homophily-guided graph convolution network which can be universally suitable for both homophilic and heterophilic networks. In specific, we incorporate a learnable homophily degree matrix into graph convolution framework for modeling the homophily and heterophily of networks and further adaptively changing the propagation process according to homophily degree between node pairs. The homophily degree matrix is learned from attribute and topology information via extracting class-aware information, which can conduct the propagation process. In return, the result of propagation process can further help learn homophily degree matrix through downstream semi-supervised tasks. These two process can be enhanced by each other and trained jointly. Experiments on seven real-world datasets demonstrate that the proposed new approach outperforms existing methods under heterophily, while gains competitive performance under homophily.

\section{Acknowledgments}
This work is supported by the National Natural Science Foundation of China (61876128, 61772361) and the George Washington University Facilitating Fund (UFF) FY21.

\bibliography{sample-base}
\end{document}